\documentclass[10pt,twocolumn,letterpaper]{article}

\usepackage{cvpr}
\usepackage{times}
\usepackage{epsfig}
\usepackage{graphicx}
\usepackage{amsmath}
\usepackage{amssymb}
\usepackage{bbm}

\usepackage[utf8]{inputenc}
\usepackage{geometry}
\usepackage{subcaption}
\usepackage[justification=centering]{caption}
\usepackage{colortbl}
\definecolor{lightgray}{rgb}{0.9,0.9,0.9}

\usepackage[T1]{fontenc}    % use 8-bit T1 fonts
\usepackage{url}            % simple URL typesetting
\usepackage{booktabs}       % professional-quality tables
\usepackage{amsfonts}       % blackboard math symbols
\usepackage{nicefrac}       % compact symbols for 1/2, etc.
\usepackage{microtype}      % microtypography

\usepackage[linesnumbered,ruled,vlined]{algorithm2e}%[ruled,vlined]{  
\usepackage{algpseudocode}  
\usepackage{amsmath}  
\usepackage[breaklinks=true,bookmarks=false]{hyperref}

\cvprfinalcopy % *** Uncomment this line for the final submission

\def\cvprPaperID{****} % *** Enter the CVPR Paper ID here

\begin{document}

%%%%%%%%% TITLE
\title{Generate Identity-Preserving Faces by Generative Adversarial Networks}
\author{
Zhigang Li\\
Department of Automation\\
Tsinghua University\\
Beijing, China 100084 \\
\texttt{lzg15@mails.tsinghua.edu.cn} \\
\and
 Yupin Luo \\
 Department of Automation\\
 Tsinghua University\\
 Beijing, China 100084 \\
 \texttt{Luo@tsinghua.edu.cn} \\
 }

\maketitle
%\thispagestyle{empty}

%%%%%%%%% ABSTRACT
\begin{abstract}
Generating identity-preserving faces aims to generate various face images keeping the same identity given a target face image. Although considerable generative models have been developed in recent years, it is still challenging to simultaneously acquire high quality of facial images and preserve the identity. Here we propose a compelling method using generative adversarial networks (GAN). Concretely, we leverage the generator of trained GAN to generate plausible faces and FaceNet as an identity-similarity discriminator to ensure the identity. Experimental results show that our method is qualified to generate both plausible and identity-preserving faces with high quality. In addition, our method provides a universal framework which can be realized in various ways by combining different face generators and identity-similarity discriminators. 
%   The ABSTRACT is to be in fully-justified italicized text, at the top
%   of the left-hand column, below the author and affiliation
%   information. Use the word ``Abstract'' as the title, in 12-point
%   Times, boldface type, centered relative to the column, initially
%   capitalized. The abstract is to be in 10-point, single-spaced type.
%   Leave two blank lines after the Abstract, then begin the main text.
%   Look at previous CVPR abstracts to get a feel for style and length.
\end{abstract}

%%%%%%%%% BODY TEXT

Generating identity-preserving faces has a wide range of applications. Considering a common scenario that the police need to search a suspect with only one picture of the front view, generating more pictures of the target person with different poses or expressions will support the task a lot. This paper focuses on generating identity-preserving faces: Given a target face image, the task is to generate various face images of the same identity with different attributes. Figure1 is the illustration. 

\begin{figure}[h]
\centering\includegraphics[width=2.5in]{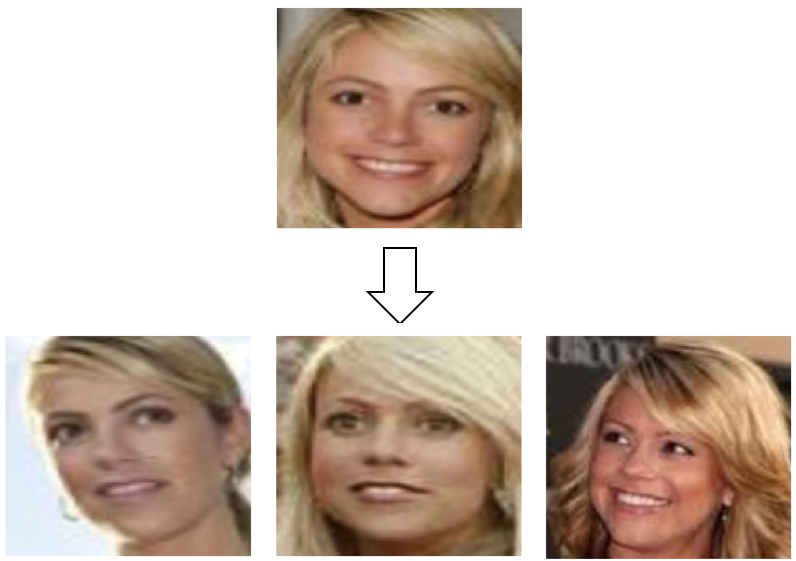}
\caption{The illustration of generating identity-preserving faces (these pictures come from LFW dataset).}
\end{figure}

The performance of generative models has been improved a lot in recent years. For instance, recent applications of GAN [1] and Variational Auto-Encoder (VAE) [2] have shown excellent capacities in generating plausible images. However, when it comes to generate identity-preserving faces, problems appear. The VAE can generate identity-preserving faces, but the generative faces tend to be blurry relative to GAN. The previous work using GAN is able to generate plausible faces, but the identity is lost. This paper is exactly to deal with such a dilemma. 

The point is how to acquire high quality of facial images and simultaneously preserve the identity. To acquire high quality of face images, we use the generator of a trained GAN model to generate faces. Then the main task is to preserve identity. We introduce a discriminator to measure the identity-similarity, thus assuring the faces we generate have the same identity with the target person. Here, we use the FaceNet [3] as the discriminator.

The idea of our method comes from daily life: most people have experienced that they mistake a stranger as an acquaintance for their similar looks. Considering that the number of people a person meets in daily life is limited, these facts indicate facial attributes are finite. Then, a face training set exists covering all types of facial attributes, which enables us to generate almost all kinds of faces. We have to state that our task is to generate identity-preserving face images that are hard to distinguish for common people but not for those who are familiar with the target person of the image.

The main contributions of our work lie in three folds: 1) We propose a pipeline to generate identity-preserving faces. 2) Experimental results show that our method is qualified to generate both plausible and identity-preserving faces with high quality. 3) The method we propose is general and can be realized in different ways.

\section{Related Work}

Many recent works have made progress in generating realistic images, where GAN and VAE stands out. GAN is good at generating plausible faces, say faces generated by GAN can scarcely be distinguished even by human [1]. The Conditional GAN [4] can generate random faces with different features by controlling the extra conditional information. Similarly, the Information Maximizing GAN (InfoGAN) [5] can control some properties of the output images by introducing the mutual information to the network. All these methods are able to generate faces with certain features, such as long hair, smile, with glasses et al. but they are hard to generate face images of the same identity with the target person. Additionally, Grigory Antipov et al. propose Age-cGAN [6] which is able to achieve identity-preserving face aging. However, Age-cGAN focuses on the age attribute while we focus on generating face images with various attributes. When it is comes to VAE models, the VAE [2] can generate faces of the same identity; the Conditional VAE [7] can control the attributes of the output faces by introducing extra vector to the input. However, the generative faces tend to be blurry relative to GAN. The Adversarial Autoencoders [8] combines GAN and VAE. It can generate clearer face images than VAE. However, with the enhancement of the image quality, the ability of preserving identity decreases.

Identity-preserving faces is somewhat similar with the style transfer. Style transfer is to output a recomposed image by transferring the reference style to the input image while faithfully preserving its content [9]. However, they are different for that generating identity-preserving faces focuses on a local region of the image, such as the open or closed mouth. By contrast, the style transfer focuses on the content of the whole picture. Changing style does not equal to changing face attributes, and vice versa. Face synthesis [10] is another similar work. Face synthesis is to convert an informal face image to a formal one, such as synthesizing frontal view faces from profiles. Differently, our task is to generate faces of various attributes with the same identity.

\section{Basics}

\subsection{Generative adversarial networks}
GAN was first proposed by Goodfellow et al. in 2014. GAN is based on the minimax game theory, implemented by a system consisting of a generator and a discriminator that compete with each other [1]. The generator generates data as real as possible to fool the discriminator while the discriminator tries its best to distinguish the generative data from those in training set. Finally, the generator can output data that are indistinguishable from those in training set for the discriminator. The process can be described as follows:

\begin{center}
  $ \min_{G}\max_{D}V(D,G)=\mathbb{E}_{x\sim p_{data}(x)}[logD(x)]+\mathbb{E}_{z\sim p_{z}(z)}[log(1-D(G(z)))] $
\end{center}

Although the performance of the GAN is well, some disadvantages, such as the well-known unstable training process and mode collapse problem, restrict the widespread use of the GAN. In 2015, Radford et al. [11] proposed DCGAN which was easy to train. Also, the DCGAN has the property of "vector arithmetic" that can be used to change specific attributes of images. We use DCGAN to generate face images, and take advantage of "vector arithmetic" to modify any interested attribute of them.

\subsection{vector arithmetic}

The "vector arithmetic" is a phenomenon of GAN (see Figure2). We represent the process by using it to modify an opened mouth face image to a closed mouth one. Input the original vector to the generator of a trained GAN model, then get the original face which is an image of a woman with mouth opened. For the same generator, the vector1 corresponds to an image of someone with mouth closed and vector2 corresponds to one with mouth opened. Then we perform arithmetic operation of ''Final vector = Original vector+vector1-vector2''. Finally, input the ''Final vector'' to the generator, we get an image of a woman with mouth closed. Additionally, other attributes of the two women keep the same, which means this arithmetic mainly change one attribute (here is the attribute of mouth opened) and almost keep all the other attributes.

Note: To make the performance of "vector arithmetic" stable, we use the average of several images for vector1 and vector2 (see Figure3). 

\begin{figure}[h]
  \centering
  \includegraphics[width=3in]{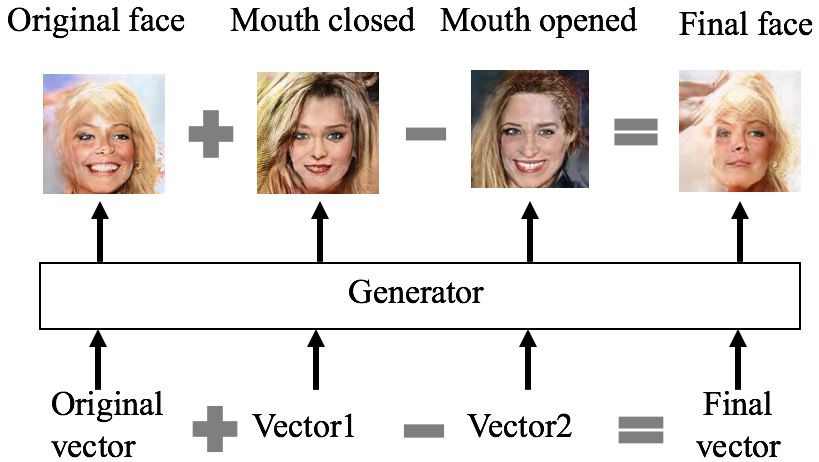}
  \caption{"vector arithmetic" of GAN}
\end{figure}

\begin{figure*}
  \includegraphics[width=\textwidth]{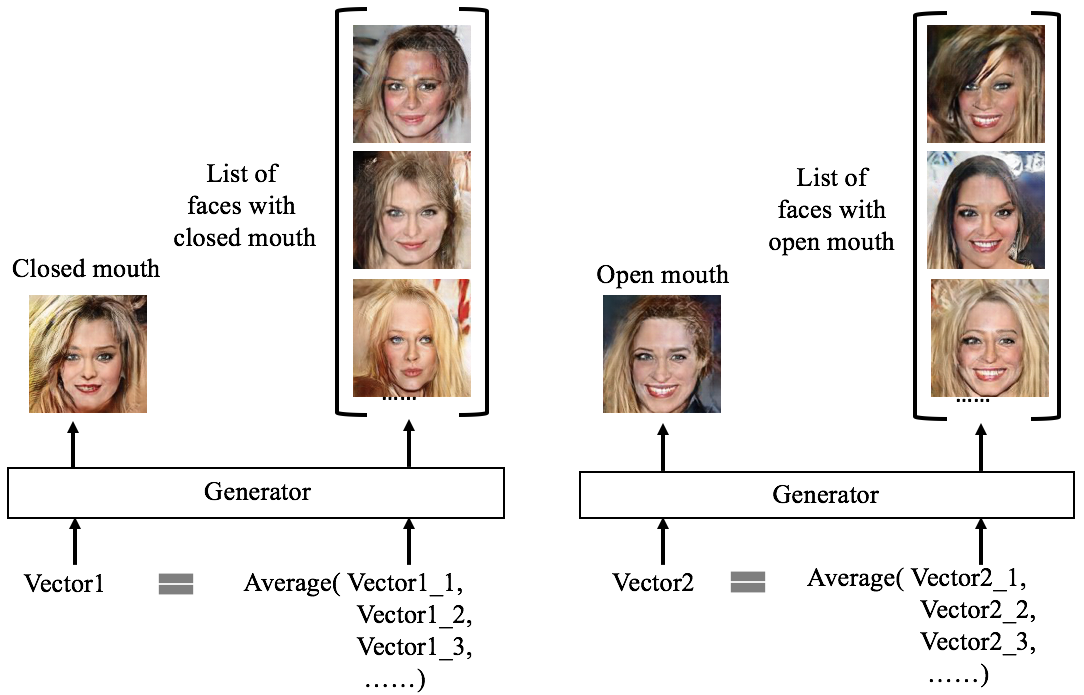}
  \caption{Method to obtain vector1 and vector2}
\end{figure*}

\subsection{FaceNet}

FaceNet [3] is a deep convolutional network which is trained using triplet loss. A well-trained FaceNet model can map face images to an embedding space where the squared L2 distances directly correspond to the similarities of these images. Figure4 shows the result of the FaceNet when inputting image pairs. 

\begin{figure}[h]
  \centering
  \includegraphics[width=3in]{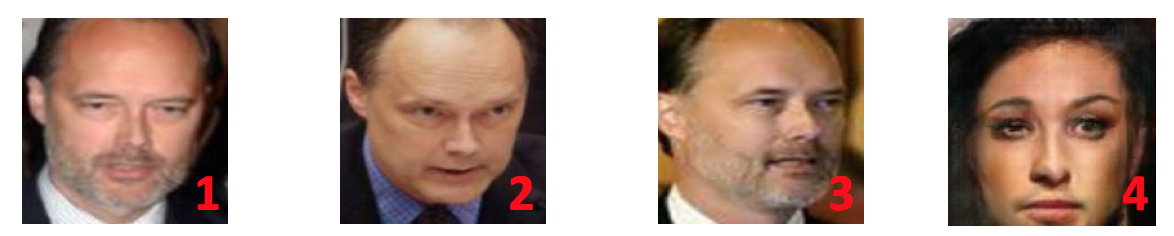}
  \caption{Visualization of samples of the input face images to FaceNet (pictures come from LFW dataset)}
\end{figure}

\section{Methods}

We generate identity-preserving faces by the following process: First, train a face generator. We use the generator of a well-trained GAN model to produce various face images; Second, select an identity-similarity metric. We choose the output of FaceNet as the metric to measure the identity-similarity between each generative face image and the target face image. Third, search the input vector whose generative face image is the most similar with the target. Finally, using "vector arithmetic" to generate various identity-preserving faces based on the face image we get at the third step. The whole process of our system is shown in Figure5.

\subsection{Train a face generator}

The image quality of face images we generate depends on the performance of generator. By using the generator of GAN as the face generator, we can generate plausible faces with high quality. 

\begin{figure*}
  \includegraphics[width=\textwidth]{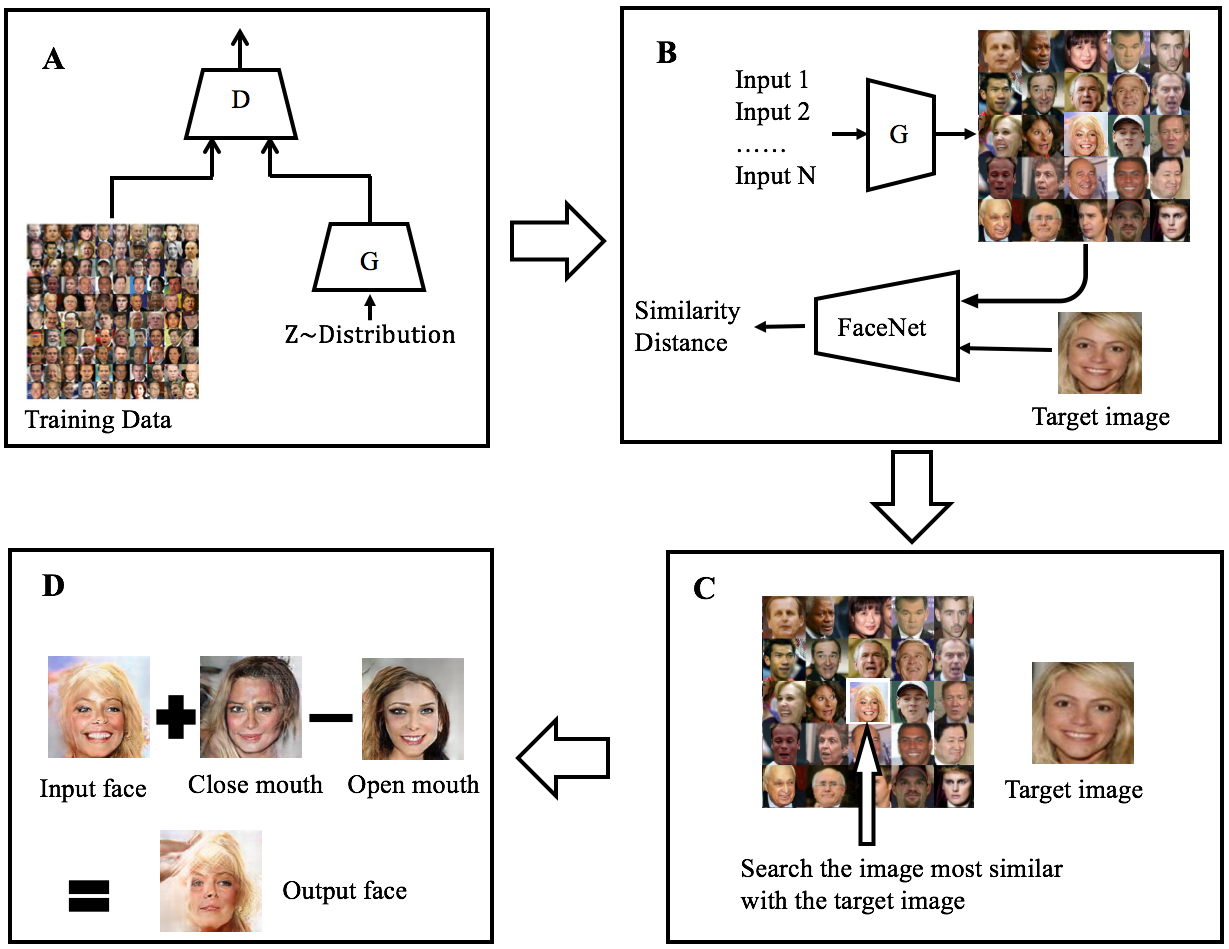}
  \caption{The process of our method}
\end{figure*}

\begin{table}
    \centering
    \begin{tabular}{|c|c|}
        \hline Face pairs & Output of FaceNet  \\ \hline
        (1,2) & 0.28174594 \\
        (1,3) & 0.60358595 \\
        (1,4) & 2.22597405 \\
	(2,3) & 0.74043938 \\  
	(2,4) & 2.18869435 \\
	(3,4) & 2.15149067  \\ \hline
    \end{tabular}
    \caption{The output of FaceNet(The lower the score is, the more similar the images are)}
    \label{tab:1}
\end{table}

\subsection{Select an identity-similarity metric}

We need an identity-similarity metric to find a generative image which has the most similar identity with the target face. Here we use the FaceNet as a discriminator to achieve the goal. Input each generative face and the target face as an image pair to the FaceNet, then it will output the identity-similarity distance score of each pair. The lower the score is, the more similar the images are (see Figure4 and Table1). 

\subsection{Search a satisfactory input vector}

Although we can use FaceNet to measure the identity-similarity of each generative face with the target, it is difficult to find the input vector whose generative face image is the most similar one with the target for the large input space. In our generator, the length of the input vector is 200 where every element ranges from -1 to 1. Then the input space ranges from -I to I, where I = [1,1,$\cdots$,1]. It can be regarded as a sphere with radius one in 200-dimension space, which is too large to traverse. To solve this problem, we propose a greedy search strategy (see Algorithm 1) to find the satisfactory input vector based on three properties of the generator of GAN. 

Our sufficient experiments show the generator of GAN owns three properties:

1) Adding small noise to the input vector does not change the identity (see Figure6.A). Note that the amplitude of the noise is not over 0.5 when that of input is 1.

2) Perform the sign function arithmetic on the input vector. Inputting the result to the generator does not change the identity (see Figure6.B).

3) Changing the amplitude of the input vector of the generator does not change the identity of the output face (see Figure6.C).

Property 1 demonstrates that the face identity keeps the same with its surroundings, so we first randomly search a temporal optimal input vector I\_opt from the global input space. Then we implement coarse search and fine tuning on I\_opt to find the satisfactory input vector. In coarse search, we change the value of alpha to control the input vectors to approach I\_opt. According to the property 2, the generative face images of these input vectors will be similar with that of I\_opt for the outputs of the sign function on them will be similar. In the fine tuning on I\_opt, we change the value of beta to add different noise to the input vectors. Note that manipulations of both coarse search and fine tuning will make the range of input vector over the initial input space [-I, I], but according to property 3, the effect is negligible.

%\begin{spacing}{1.0}

\begin{algorithm}  
  \caption{Search a satisfactory input vector}  
  \KwIn{The generator of the trained GAN model; Trained FaceNet; Target face image I\_target(Note: we use FaceNet(im1, im2) to indicate the output of FaceNet when the input image pair is im1 and im2)  }  
  \KwOut{A satisfactory input vector I\_opt}  

$\backslash\backslash$Initialization

Set N = 1000, I = [1,1,$\cdots$, 1]$_{200}$, input space S = [-I, I], Similarity threshold T = 0.4

$\backslash\backslash$Get an initial I\_opt

Sample N input vectors  from the S randomly, denoted as V.

I\_opt = Select\_Optimal (V, I\_target)

$\backslash\backslash$Stage one: coarse search

 \While{FaceNet(I\_opt , I\_target) $>$ T} 
 {
	Sample  input vectors  from the [-I+$\alpha$$\cdot$I\_opt, I+$\alpha$$\cdot$I\_opt] randomly.
	
 	 I\_opt  = Select\_Optimal (V, I\_target)
	
 	\If{$\alpha$ $<$ 1:}
 	{
		$\alpha$  = $\alpha$ +0.1
 	}
 }

$\backslash\backslash$Stage two: fine tuning

\While{FaceNet(I\_opt , I\_target) $>$ T} 
{
	Sample  input vectors  from the [-I\_opt+$\beta$$\cdot$I, I\_opt+$\beta$$\cdot$I] randomly.

	 I\_opt  = Select\_Optimal (V, I\_target)
	
 	\If{$\alpha$ $<$ 1:}
 	{
		$\alpha$  = $\alpha$ +0.1
 	}
 }
 
 return I\_opt 

$\backslash\backslash$Subfunction to chose the optimal input vector from the set V

Select\_Optimal (V, I\_target)

I\_opt  = V [1] 

 \For{I in V}  
 {  
    \If{FaceNet(I, I\_target) $<$ FaceNet(I\_opt , I\_target) }  
    {  
        I\_opt  = I
    }  
 } 
 return I\_opt 

\end{algorithm}  

%\end{spacing}

\begin{figure}
  \centering
  \includegraphics[width=3in]{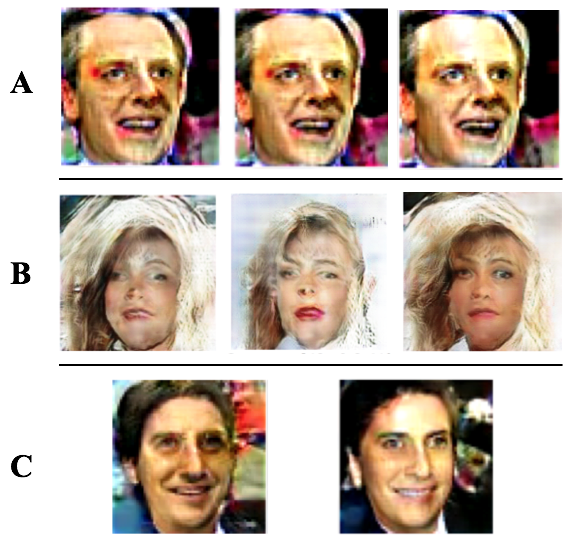}
  \caption{The illustration of the three properties of the generator of GAN. A shows the effect of adding small amplitude noise to the input (amplitude of noise is less than 0.5 while input is 1), cconsequently the identity attribute dose not change. B shows the relation between the sign of input with the identity attribute. The identity attribute changes a little even using the sign of input to generate face. C shows the relation between the amplitude of input with the identity attribute. Change the amplitude of the input, the identity attribute dose not change}
\end{figure}

\begin{figure*}
  \centering
  \includegraphics[width=4.7in]{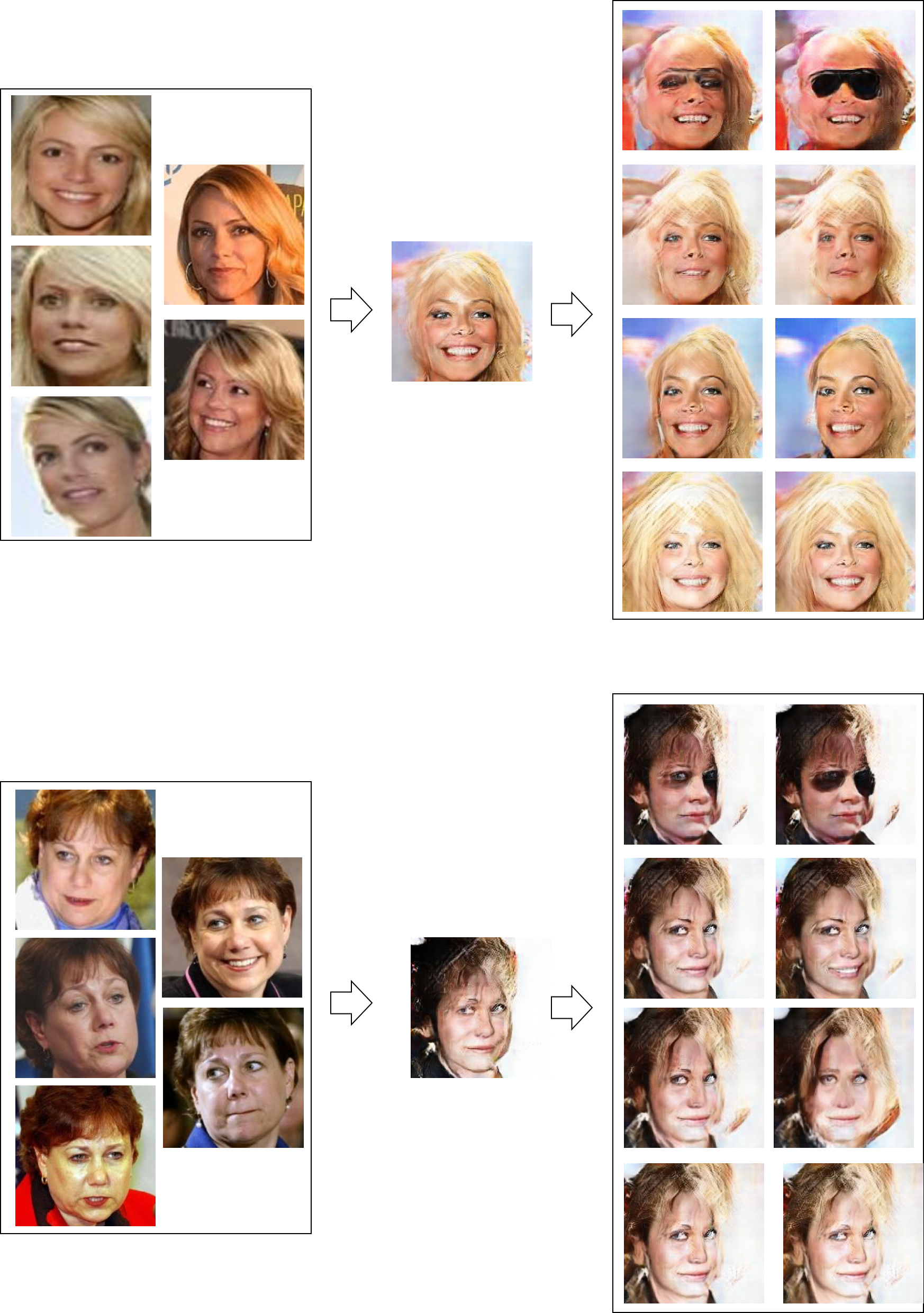}
  \caption{Visualization of identity-preserving face image generated by our model. The left are target images. The middle image is generated and found by the greedy search strategy . The right are generative identity-preserving faces using "vector arithmetic"  with various attributes: with or without glasses, closed or open mouth, face right or left, dark or bright.}
\end{figure*}

\subsection{Generate identity-preserving faces with various attributes}

Once getting the satisfactory input vector which can generate an identity-similarity face with the target, we can use many methods to generate various identity-preserving faces with different attributes. The methods include: changing the amplitude of the input vector, adding noise to the input, using the "vector arithmetic", etc. Here, we use "vector arithmetic".

\subsection{Generalization}

Our method can be generalized as a universal one. The face generator can be any one only if it can generate various faces, and the discriminator can be also any one only if it can measure the degree of the identity-similarity. Then, we generalize our method: First, train a face generator to generate various face images; Second, choose an identity-similarity metric to test the generative faces with the target; Third, search the input vector whose generative face image is the most similar with the target; Finally, generate more various identity-preserving faces based on the input vector we find at the third step. By choosing different face generator and identity-similarity metrics, we can get different framework to generate identity-preserving face images.

\section{Experiment}

To ensure that the training set consists of most facial features, we choose CelebA database [12] as the training set to meet the demand of diversity. The target face images are come from LFW database [13]. Table2 shows the details of these databases.

\begin{table}[h]
    \centering
    \begin{tabular}{|c|c|c|}
        \hline  Database& \#Subjects & \#Images \\ \hline
        LFW & 5749 & 13233 \\
        CelebA & 10177  & 202599 \\ \hline
    \end{tabular}
    \caption{LFW and CelebA Databases}
    \label{2}
\end{table}

We choose five face images of the target person from the LWF database and google search engine(see Figure7, left), which means these images are not in the training set. When measuring the similarity between each generative face image and the target, we actually compute five identity-similarity distance scores by FaceNet and use the average as the final score. The identity-preserving face images generated and found by the greedy search strategy are shown in the Figure7, middle. Using "vector arithmetic", we generate more identity-preserving face images with different attributes (see Figure7, right).

We still use the FaceNet to evaluate the performance of our method. Results show that the distance score of the two experiments between the generative face image and the target are separately 0.64 and 0.60(see Table3). Note that the distance score between two real images of the same person reaches 0.74 (see Table1), therefore, our method is effective. 

\section{Conclusion}

We fusion the generator of GAN and the FaceNet to generate high quality identity-preserving face images. The experiments show our method is effective to generate images without blurry problem. Additionally, our method provides a universal framework, which can be realized in different ways by choosing different generative models and discriminators. To the best of our knowledge, this is the first time to use GAN to generate identity-preserving faces.  

\section{Acknowledgments}

We would like to give the deepest appreciation to Jie Wang for her patient discussions and great help on this paper. 

\bibliographystyle{model1-num-names}

\begin{thebibliography}{00}

\bibitem{key} Goodfellow I, Pouget-Abadie J, Mirza M, et al. Generative adversarial nets[C]//Advances in neural information processing systems. 2014: 2672-2680.

\bibitem{key} Kingma D P, Welling M. Auto-encoding variational bayes[J]. arXiv preprint arXiv:1312.6114, 2013.

\bibitem{key} Schroff F, Kalenichenko D, Philbin J. Facenet: A unified embedding for face recognition and clustering[C]//Proceedings of the IEEE Conference on Computer Vision and Pattern Recognition. 2015: 815-823.

\bibitem{key} Gauthier J. Conditional generative adversarial nets for convolutional face generation[J]. Class Project for Stanford CS231N: Convolutional Neural Networks for Visual Recognition, Winter semester, 2014, 2014(5): 2.

\bibitem{key} Chen X., Duan Y., Houthooft R. et al. Infogan: Interpretable representation learning by information maximizing generative adversarial nets[C]//Advances in Neural Information Processing Systems. 2016: 2172-2180.

\bibitem{key}  Antipov G, Baccouche M, Dugelay J L. Face Aging With Conditional Generative Adversarial Networks[J]. arXiv preprint arXiv:1702.01983, 2017.

\bibitem{key} Yan X, Yang J, Sohn K, et al. Attribute2image: Conditional image generation from visual attributes[C]//European Conference on Computer Vision. Springer International Publishing, 2016: 776-791.

\bibitem{key} Makhzani A, Shlens J, Jaitly N, et al. Adversarial autoencoders[J]. arXiv preprint arXiv:1511.05644, 2015.

\bibitem{key} Jing Y, Yang Y, Feng Z, et al. Neural Style Transfer: A Review[J]. arXiv preprint arXiv:1705.04058, 2017.

\bibitem{key} Huang R, Zhang S, Li T, et al. Beyond Face Rotation: Global and Local Perception GAN for Photorealistic and Identity Preserving Frontal View Synthesis[J]. arXiv preprint arXiv:1704.04086, 2017.

\bibitem{key} Radford A, Metz L, Chintala S. Unsupervised representation learning with deep convolutional generative adversarial networks[J]. arXiv preprint arXiv:1511.06434, 2015.

\bibitem{key} Liu Z, Luo P, Wang X, et al. Deep learning face attributes in the wild[C]//Proceedings of the IEEE International Conference on Computer Vision. 2015: 3730-3738.

\bibitem{key} Huang G, Mattar M, Lee H, et al. Learning to align from scratch[C]//Advances in Neural Information Processing Systems. 2012: 764-772.

\end{thebibliography}

\end{document}